\documentclass[twoside]{article}
\usepackage{aistats2e}

\usepackage{amsmath}
\usepackage{algorithm}
\usepackage{algorithmic}
\usepackage{graphicx}
\usepackage{wrapfig} 
\usepackage{subfigure} 
\usepackage{amsfonts}
\usepackage{natbib}

\begin{document}

%

%

\twocolumn[

\aistatstitle{Texture Modeling with Convolutional Spike-and-Slab RBMs and Deep Extensions}

\aistatsauthor{ Heng Luo \And Pierre Luc  Carrier \And Aaron  Courville \And Yoshua  Bengio}

\aistatsaddress{ \texttt{\{heng.luo, pierre-luc.carrier, aaron.courville, yoshua.bengio\}@umontreal.ca}  
\\Department of Computer Science and Operations Research
\\University of Montreal
} ]

\begin{abstract}
We apply the spike-and-slab Restricted Boltzmann Machine (ssRBM) to
texture modeling. The ssRBM with
tiled-convolution weight sharing (TssRBM) achieves or surpasses the state-of-the-art 
on texture synthesis and inpainting by parametric models.  
We also develop a novel RBM
model with a spike-and-slab visible layer and binary variables in the
hidden layer. This model is designed to be stacked on top of the TssRBM.
We show the resulting deep belief network
(DBN) is a powerful generative model that improves on single-layer models
and is capable of modeling not only single high-resolution and challenging
textures but also multiple textures.
\end{abstract}

\vspace*{-2mm}
\section{Introduction}
\vspace*{-2mm}

Texture processing is one of the essential components of scene understanding in
human vision. Natural images can be seen as a large mixture of
heterogeneous textures. Thus, to a certain extent, progress in modeling
natural images requires that we make progress in modeling textures. To this
end, texture modeling has been an active research area of machine learning,
computer vision and graphics during the past five decades. Although
nonparametric approaches~\citep{Lin-CVPR2006}
have made significant progress in synthesizing textures from example
images, capturing the statistical properties of textures via a
probabilistic model remains an active area of inquiry.  Such probabilistic
models are important for modeling natural images~\citep{Heess2009} but also
for understanding human vision~\citep{Zhu2000}.

In this work we consider a probabilistic model of textures based on the
spike-and-slab Restricted Boltzmann Machine (ssRBM)
\citep{Courville2011a,Courville+al-2011-small}. The ssRBM has previously
demonstrated the ability to generate samples of small natural images that
preserved much of their statistical
structure~\citep{Courville+al-2011-small}. This would suggest that the
ssRBM is potentially well suited to the task of texture modeling. Following
the recent exploration of Boltzmann machines for textures
\citep{Kivinen2012}, we have trained ssRBMs with tiled-convolution weight
sharing~\citet{Gregor+LeCun-2010,Le2010} on the Brodatz-texture
images\footnote{\texttt{www.ux.uis.no/$\tilde{~}$tranden/brodatz.html}}. Tiled
convolution allows weight sharing in filters with non-overlapping receptive
fields. The use of tiled convolution is a particularly appropriate choice
of model architecture in the context of texture modeling. The weight
sharing allows the model to synthesize texture patches of variable size,
while the tiling pattern of weight sharing allows us to efficiently devote
model capacity to modeling the local texture patches.

In \citet{Kivinen2012}, the authors concentrated their quantitative
evaluation of the texture models on a subset of the Brodatz textures that
exhibit strong spatial invariance, i.e. textures largely consisting of a
regular repeating pattern. While this is an important problem in its own
right, most natural textures (i.e. those associated with a natural-looking
world) exhibit significant spatial non-stationarity and features with a
wide spatial frequency range.  One popular way to deal with images with a
wide spatial frequency range is to decompose the frequencies using, for
example, a Laplacian image pyramid. However, since many textures have
features that interact across spatial resolutions, spatial pyramids would
appear to be inappropriate.  We propose that \emph{deep} convolutional
generative architectures are well suited to model these natural
textures. In particular, by increasing the effective receptive field with
depth, we can use higher layers of the model to efficiently communicate
information such as phase to spatially isolated parts of the first
layer model.

Deep belief networks also have another important property that we find
useful in the context of texture modeling. As argued by \citet{Hinton-ipam2012,LeRoux-Bengio-2008}, 
training the lower layers by contrastive divergence
(CD)~\citep{Hinton06} allows the lower layers to concentrate on modeling local
features of the data. We have found best results by training the lower layers by CD and the
uppermost layer by a closer approximation to maximum likelihood, such as
persistent contrastive divergence (PCD)~\cite{Tieleman08}, promoting a better
division of labour between the layers of the DBN. On this account, the
ssRBM offers an important advantage over other similar models in the
literature. For unlike models such as the mcRBM~\citep{Ranzato2010b} and
mPoT~\citep{ranzato+mnih+hinton:2010}, the structure of the ssRBM makes it
readily amenable to CD training.

Our contributions are, first, the exploration of a tiled-convolutionally
trained ssRBM (TssRBM) texture model and its objective comparison with the
other similar models in the literature. We show that the TssRBM is
competitive with the state-of-the-art on texture synthesis and inpainting
tasks on a selection of Brodatz textures.  Second, we develop a novel RBM
model with a spike-and-slab visible layer and binary variables in the
hidden layer. This model is design to be stacked on top of the TssRBM
within a deep belief network configuration, with each layer trained
convolutionally with a greedy layer-wise pretraining strategy. We
demonstrate how the resulting two and three-layer DBN (the third layer is a
standard RBM) models are able to encode longer term dependencies in the
higher layers while simultaneously recovering more detailed structure in
the CD-trained lower layer, all of which translates to superior texture
model performance -- particularly when the textures being modeled exhibit
strong non-stationarity.  Finally, we show how the depth helps in learning
a generative model of multiple textures. \citet{Kivinen2012} introduce a
model capable of modeling multiple textures, however they make use of label
information in the training process alleviating the difficult learning
problem of constructing multiple modes to represent each texture. In this
work, we show how a deep belief network based on the ssRBM is capable of
learning to model multiple textures based on purely unsupervised training.

\vspace*{-2mm}
\section{Previous Work}
\vspace*{-2mm}


The problem of texture synthesis has been extensively studied in the
computer vision community for decades \citep{Zhu2000}. Probably the most
popular texture synthesis strategies are currently example-based or
nonparametric methods \citep{Wei2009}. These typically seed a target image
with transformed versions of patches drawn from the target texture. While
these methods are flexible, they are unlikely to be readily applicable to
natural textures, 
where some aspects of the statistical
structure (eg. the path of the duck tracks) are global in scope.

The Gaussian RBM ~\citep{Welling05,Ranzato2010b} models real-valued observations by adding quadratic terms
on the visible units to the standard binary-binary RBM energy function.
One limitation of the Gaussian RBM is that changing its hidden unit activations only 
changes the conditional mean of the visible units. For modeling natural images, it has
been found important to allow the hidden unit configuration to capture
changes in the covariance between pixels, and this has motivated
several of the models discussed below as well as the ssRBM.
The product of Student's T-distributions (PoT) model~\citep{Welling2003a-small}
is an energy-based model where the conditional distribution over the
visible units conditioned on the hidden variables is a multivariate
Gaussian (non-diagonal covariance) and the complementary conditional
distribution over the hidden variables given the visibles are a set of
independent Gamma distributions. The PoT model has recently been
generalized to the mPoT model~\citep{ranzato+mnih+hinton:2010-short} to
include nonzero Gaussian means by the addition of Gaussian RBM-like hidden
units. In the same work, \citet{Kivinen2012} explored the
``Multi-Texture Boltzmann Machine'' (Multi-Tm), training a single large Gaussian
RBM (up to 256 feature maps par tiling position, as opposed to 32
maps per tiling position) on multiple textures. In modeling multiple textures, 
\citet{Kivinen2012} used label information during the training process to enable 
the model to focus on a single texture class at a time. In section \ref{sec:multi_texture}, 
we show how we can use a deep belief network, based on the ssRBM, to learn a model of 
multiple textures using no label information at all.

In addition to validating the ssRBM as the basis of an effective texture
model, we also set out to study the impact of adding layers to
the tiled-convolutional ssRBM model, in order to see if depth can
help maintain coherence of large scale texture features. Recent work
\citep{ranzato:on} has shown that stacking additional RBM layers on top of an mPoT
model (also trained using tiled-convolutional weight sharing) can have a
dramatic impact of the ability of the model to generate globally coherent
natural image samples. Findings such as these motivated our attempt to use
depth to synthesize textures with increased global coherence.

\vspace*{-2mm}
\section{Spike-and-Slab RBM}
\label{sec:TssRBM}
\vspace*{-2mm}

The ssRBM describes the interaction between three sets of random variables:
the real-valued visible random vector $v\in\mathbb{R}^{D}$ representing the
observed data of dimension $D$, the set of binary ``spike'' random
variables $h \in [0,1]^{N}$ and the real-valued ``slab'' random variables
$s\in\mathbb{R}^{N}$.
The ssRBM has the interpretation that, with $N$ hidden units, the $i$th
hidden unit is associated with \emph{both} an element $h_i$ of the binary
vector and an element $s_i$ of the real-valued variable.
In this work we will concern ourselves with the ssRBM formulation referred
to as the $\mu$-ssRBM \citep{Courville+al-2011-small} with the associated
energy function:

{\small
\begin{multline}
\vspace*{-2mm}
\textstyle
E(v,s,h)=-\sum_{i=1}^{N}v^{T}W_{i}s_{i}h_{i}\
+\frac{1}{2}v^{T}\left(\Lambda+\sum_{i=1}^{N}\Phi_{i}h_{i}\right)v \\
+\frac{1}{2}\sum_{i=1}^{N}\alpha_{i}s_{i}^{2}\
-\sum_{i=1}^{N}\alpha_{i}\mu_{i}s_{i}h_{i}\ -\sum_{i=1}^{N}b_{i}h_{i} +\frac{1}{2}\sum_{i=1}^{N}\alpha_{i}\mu_{i}^2h_{i},
\label{eq:mu_energy}
\end{multline}
}
where $W_{i} \in \mathbb{R}^{D}$ denotes the $i$th weight (or feature)
vector, $b_i$ is a scalar bias associated with the spike variable $h_i$,
$\mu_i$ and $\alpha_i$ are respectively a mean and precision parameter
associated with the random slab variable $s_i$, $\Lambda$ is a diagonal precision matrix on the
visibles $v$, and $\Phi_i$ is an $h_i$-gated contribution to the precision
on $v$.  As is standard with energy-based models, the joint probability
distribution over $v$, $s = [s_1,\dots,s_N]$ and $h = [h_1,\dots,h_N]$ is
specified as: $p(v,s,h)=\frac{1}{Z}\exp\left\{ -E(v,s,h)\right\}$,
where $Z$ is the normalizing partition function. 

An interesting property of the ssRBM is that despite having
higher-order interactions of variables, the model maintains the bipartite
graph structure of the standard restricted Boltzmann machine where
the $i$th hidden unit consists of the product of the random variables $s_i$
and $h_i$. This property implies that, unlike the mPoT (which
also models conditional variance), the ssRBM shares the simple and practical
conditional independence structure of the standard restricted Boltzmann
machines. This makes it easy to use efficient block Gibbs sampling. 
As seen in the model conditionals: 

\vspace*{-9mm}
{\small
\begin{eqnarray}
p(v\mid s,h) & = &  \mathcal{N}\left(C_{v \mid s, h}\sum_{i=1}^{N}W_{i}s_{i}h_{i}
  \ ,\ C_{v \mid s, h}\right) \label{eq:mu_p(v|s,h)}, \\
P(h \mid v) & = & \prod_{i=1}^{N} \sigma\left(\frac{1}{2}\alpha_{i}^{-1}(v^{T}W_{i})^{2}+v^{T}W_{i}\mu_{i}\right.\\
 & & \;\;\;\; \left.-\frac{1}{2}v^{T}\Phi_{i}v+b_{i}\right), \label{eq:mu_p(h|v)a} \\ 
p(s \mid v,h) & = & \prod_{i=1}^{N}\mathcal{N}\left(\left(\alpha_{i}^{-1}v^{T}W_{i}+\mu_{i}\right) h_{i}\ ,\ \alpha_{i}^{-1}\right), \label{eq:mu_p(s|v,h)}
\end{eqnarray}
}
\vspace*{-5mm}

where $\mathcal{N}(\mu,C)$ denotes a Gaussian distribution with mean
$\mu$ and covariance $C$, $\sigma$ represents a logistic sigmoid, and
$C_{v \mid s, h} = \left(\Lambda+\sum_{i=1}^{N}\Phi_{i}h_{i}\right)^{-1}$
is the diagonal conditional covariance matrix.

\paragraph{Training the ssRBM:} Like the standard RBM, learning and inference in the ssRBM is rooted in the
ability to efficiently draw samples from the model via block Gibbs
sampling. In training the ssRBM we are free to use either contrastive divergence or a better approximation to maximum likelihood such as the stochastic maximum 
likelihood algorithm, also known as
persistent contrastive divergence (PCD)~\citep{Tieleman08}.
CD training involves approximating the negative phase component of the likelihood gradient by 
a few steps (often just one) of Gibbs sampling away from the data presented in the positive phase. 
In PCD, one maintains a persistent Markov chain to approximate the negative phase and simulates a few Gibbs steps between each parameter update. These samples are then used to approximate the
expectations over the model distribution $p(v,s,h)$. Details regarding PCD training of the ssRBM are available in \cite{Courville2011a}.

Our use of block Gibbs sampling marks an important distinction between our
approach to learning and that used by \citet{Kivinen2012}, who use Hybrid
Monte Carlo (HMC) \citep{Neal94} to draw samples from the model
distribution. Their use of HMC is likely motivated by their need to train
models such as the PoT model where the conditional over visible vectors do
not factorize and hence is not amenable to efficient block Gibbs sampling. The ability to easily and efficiently Gibbs sample from the ssRBM also makes it amenable to CD training, unlike models such as the PoT and mPoT models. As we show in the experiments with deeper models, the use of CD training is crucial to achieving our best results.


\paragraph{Tiled-Convolutional ssRBM:}
\citet{Gregor+LeCun-2010} introduced tiled-convolutional weight
sharing~\citep{ranzato+mnih+hinton:2010,Le2010} and is similar to convolutional
weight sharing~\citep{LeCun98-small,Desjardins-2008-small,HonglakL2009} 
except that spatially neighboring features (with overlapping
receptive fields) do not share weights. Within the tiled-convolutional structure,
every specific filter ties the input images without overlaps with itself and at
the same time different filters do overlap with each other. The first setting of 
tiled-convolution not only allows us to efficiently work on much bigger images 
than traditional convolutional models but also makes the states of hidden units less correlated, which is 
very helpful when we draw samples from the models by block
Gibbs sampling. The second setting aims at removing the tiling artifacts introduced
the non-overlapping filters.

To make comparisons easier, our TssRBM uses the same architecture as ~\citet{Kivinen2012} 
including the same receptive field size of $11\times11$ and the same diagonal tiling pattern 
with a stride of one pixel (neighboring receptive fields are offset by one pixel). This diagonal 
tiling (which reduces considerably the number of free parameters) makes for 11 sets of filters 
(one for each offset). We also kept constant the number of filters (32 per set) to 
make comparisons with the results in ~\citet{Kivinen2012} simpler.

\vspace*{-2mm}
\section{An ssRBM-based Deep Belief Network}
\label{sec:newRBM}
\vspace*{-2mm}

In this section, we describe how we extend the TssRBM in a hierarchical generative model in the form of a deep belief network (DBN). Following the standard procedure for learning DBNs, we follow a layer-wise training strategy. Training the bottom layer ssRBM, either by CD, PCD or FPCD is straightforward and discussed above in Sec. \ref{sec:TssRBM}. We now consider the form of the model we intend to stack on top of the ssRBM.

Following the DBN approach, we express the ssRBM model as
\begin{equation*}
\vspace*{-1mm}
P_{\mathrm{ssRBM}}(v) = \sum_{s,h}P(v|s,h)P(s,h).
\vspace*{-1mm}
\end{equation*}
As discussed in the previous section, due to the factorial nature of $P(v|s,h)$, it is convenient
to consider this the bottom layer of our DBN and focus on how to model the spike-and-slab latent state. Let $P^0(v)$ denote the data distribution. We introduce
another higher-layer model of the spike-and-slab state $Q(s,h)$ to model the aggregated posterior 
distribution, $Q^0(s,h)$, of the ssRBM
\begin{equation*}
\vspace*{-1mm}
Q^0(s,h) = \sum_{v}P(s,h|v)P^0(v)
\vspace*{-1mm}
\end{equation*}
If $Q(s,h)$ models the aggregated posterior $Q^0(s,h)$ better than does
$P(s,h)$ (defined by the ssRBM), then adding the second layer can improve
the model of the training data~\citep{Hinton06}.

Formally, the two layer model is,
\begin{equation*}
\vspace*{-1mm}
P_{2-\mathrm{layer}}(v) = \sum_{s,h}P(v|s,h)Q(s,h)
\vspace*{-1mm}
\end{equation*}
From a generative perspective, the sampling procedure consists of generating a sampling pair $(s,h)$ from the top (second here) layer, followed by mapping them to image space though $P(v|s,h)$.

We have yet to specify the form of the model $Q(s,h)$. We will follow the common practice of using another model of the RBM family to model the distribution over $s$ and $h$. We introduce a variant of the RBM: $P(s,h,g)$, which models the aggregate posterior $Q^0(s,h)$ through a hidden a binary random vector: $g \in \{0,1\}^M$.  We choose to use a binary hidden layer in order to transition to a more standard binary representation. When we include a third layer to the DBN, then that layer will be formed by training a standard binary-binary RBM.

The energy function of the second layer model is defined as follows:
\begin{multline}
\vspace*{-1mm}
E(s,h,g) = 
         - \sum_{i=1}^{N}\sum_{j=1}^{M}g_{j}U_{ij}s_{i}h_{i}
         - \sum_{j=1}^{M}\rho_{j}g_{j} \\
         + \frac{1}{2}\sum_{i=1}^{N}\alpha_{i}s_{i}^{2} 
         - \sum_{i=1}^{N}\alpha_{i}\mu_{i}s_{i}h_{i} 
         -\sum_{i=1}^{N}b_{i}h_{i}
\vspace*{-1mm}
\end{multline}
where $U_{ij}$ refers to the $ij$th element of the weight matrix encoding the interactions between 
$g_{j}$ and spike-and-slab variables $h_i$ and $s_{i}$ respectively. The term $\rho_{j}$ controls the bias on the binary $g_{j}$. All other parameters have the same interpretation as their first layer analogues. 

Similar to the standard ssRBM, the conditionals 
\mbox{$P(h \mid g)$}, \mbox{$P(s \mid h,g)$} and $p(g \mid s,h)$ are factorial and given by:
{ \small
\begin{eqnarray}
\vspace*{-1mm}
P(h_j=1 \mid g)\hspace*{-3mm}   & = &\hspace*{-3mm} \sigma\left(\frac{1}{2}\alpha_{i}\left(\alpha_{i}^{-1}\sum_{j=1}^{D}v_{j}W_{ij}+\mu_{i}\right)^{2}+b_{i}-\frac{1}{2}\alpha_{i}\mu_{i}^{2}\right) \nonumber \\
p(s \mid v,h)\hspace*{-3mm} & = &\hspace*{-3mm} \mathcal{N}\left(\left(\alpha_{i}^{-1}\sum_{j=1}^{M}v_{j}W_{ij}+\mu_{i}\right)h_{i}\ ,\ \alpha_{i}^{-1}\right) \nonumber \\
p(g_j=1 \mid s,h)\hspace*{-3mm} & = &\hspace*{-3mm} \sigma\left(\sum_{i=1}^{N}U_{ij}s_{i}h_{i}+\rho_{j}\right)
\vspace*{-1mm}
\end{eqnarray}
}

The structure of this model gives us two 
advantages. First, at the start of training the second layer we can make $Q(s,h)$ close to $P(s,h)$ defined
by the first layer ssRBM by initializing the corresponding parameters to match their first layer analogues' values.
Second, after training the second layer, we get a new binary representation for training data.
Based on it, building a even deeper model is straightforward. In our experiments, this architecture works very well.

\begin{figure}
    \centering
    \label{fig:conv_model}
    \vspace{-0.8cm}    
    \includegraphics[scale=0.4]{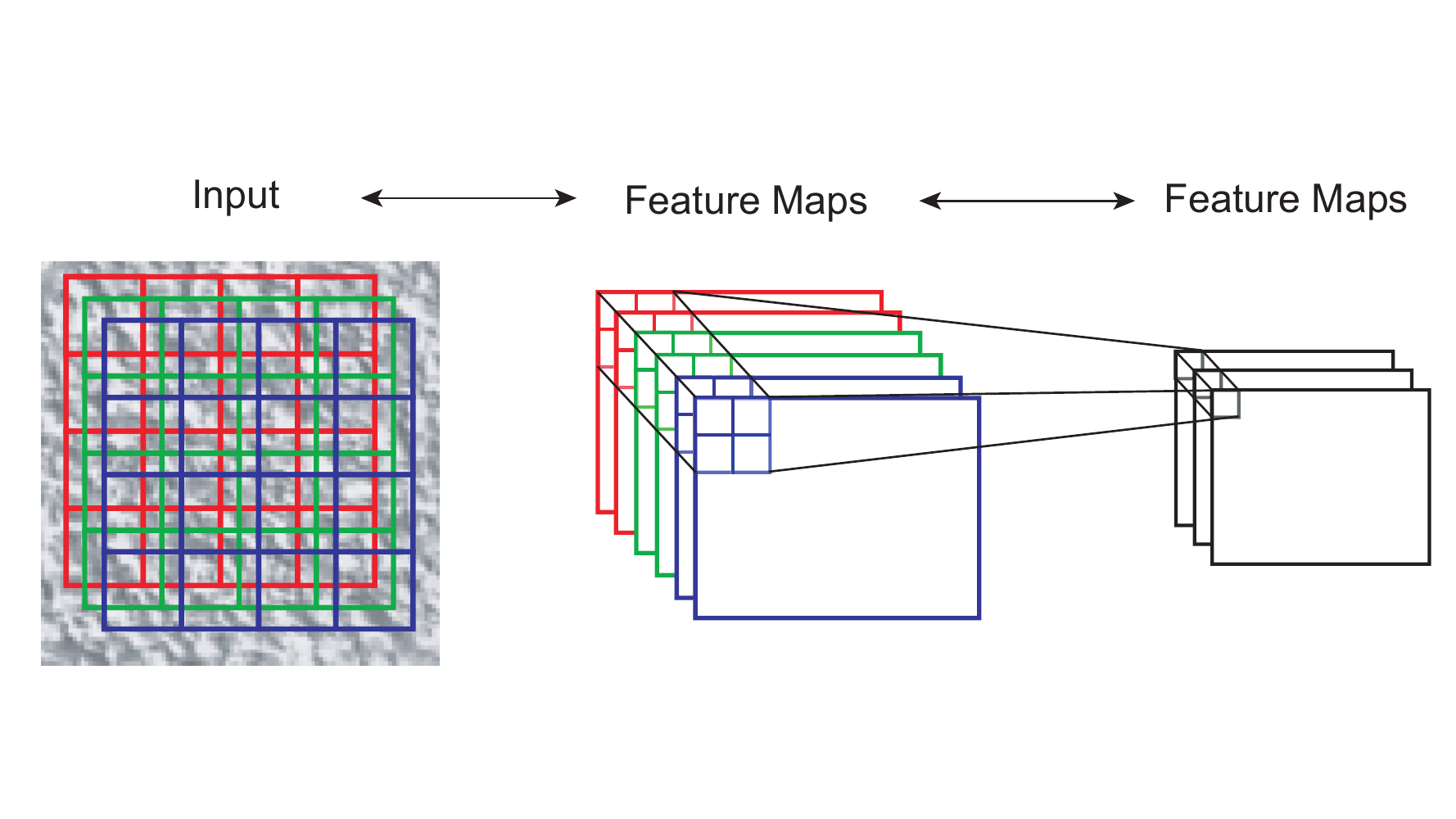}
    \vspace{-0.8cm}    
    \caption[]{\small{ 
    The architecture of the lowest two layer. The first layer possesses tiled-convolutional weight sharing in a diagonal arrangement (tilings are represented by different colors). Each second layer unit has a $2\times 2$ receptive field over all the feature maps in the first hidden layer. The second layer is arranged with traditional convolutional weight sharing and a stride of 1.}
    }    
    \vspace{-5mm}    
\end{figure}

\paragraph{Training the second-layer model:}
After pretraining the first layer (ssRBM), given training data $v$, we sample $\hat{h}$ from $p(h \mid v)$, then 
take $\mathbb{E}_{p(s \mid v,\hat{h})}[s]$ and $p(h \mid v)$ as the new training data to train the second layer. Just as we do for the bottom-layer ssRBM, we train this second-layer RBM with either PCD or with CD. We typically see best results if we train with PCD for the top-layer model and with CD for all other layers. 

\paragraph{Sampling and inference in our two layer model:}
Once the second layer has been trained with PCD it can be used to generate samples. We run Gibbs sampling in the top layer, getting the sample $\hat{g}$. Next, we sample $\hat{h}$ from $P(h \mid \hat{g})$ then pass $P(h \mid \hat{g})$ and $\mathbb{E}_{p(s \mid \hat{g},\hat{h})}[s]$
to the first layer. Inference in our two layer model is exactly the same to the process of converting training data into the new 
representation (spike and slab variables) discussed above. Given $v$, we sample $\hat{h}$ from $p(h \mid v)$, then pass
$\mathbb{E}_{p(s \mid v,\hat{h})}[s]$ and $p(h \mid v)$ to higher layer. 

\paragraph{Convolutional Structure:} The second layer possesses a convolutional weight sharing structure (not tiled-convolutional). 
Based on our use of patches of size $98 \times 98$ randomly cropped from the texture images,
the tiling structure of the first layer model results in a set of $32 \times 11$ feature maps 
of size $8 \times 8$ (the receptive field size was $11 \times 11$). Second layer hidden
units are each connected to all $32 \times 11$ feature maps with the same
$2 \times 2$ receptive field across all feature maps. Using a stride length of 1, this implies that 
each second layer feature is associated with a feature map of size $7 \times 7$. For our experiments with a 3-layer model,
we keep the same convolutional weight sharing structure for the third layer and use receptive fields of size $2 \times 2$.

\begin{figure*}[htbp!]
\centering
\includegraphics[scale=0.5]{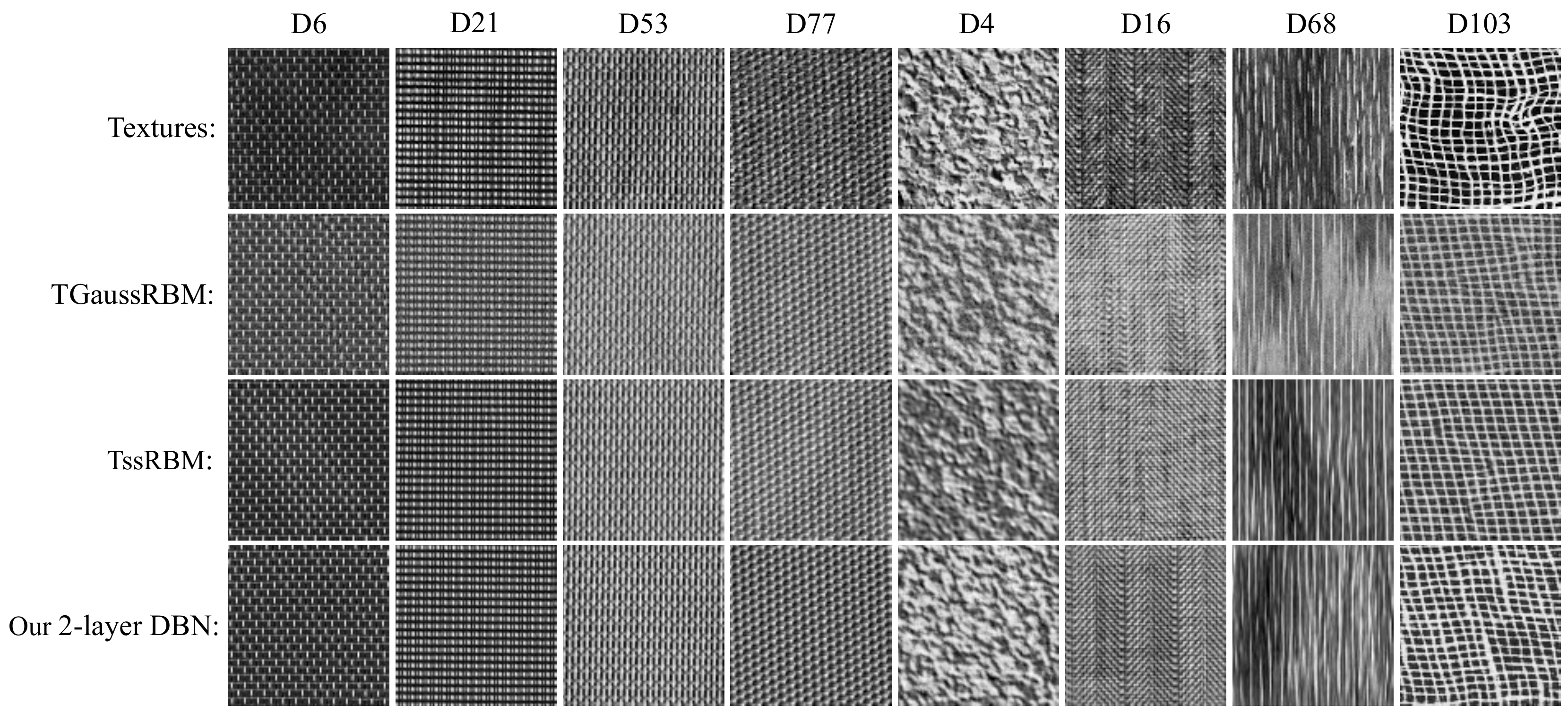}
\caption{\small Examples of texture synthesis for the models under consideration (rows) for different textures (columns).
The top row has original data.}
\label{fig:texture_synthesis}
\end{figure*}

\vspace*{-2mm}
\section{Experiments}
\vspace*{-2mm}

We evaluate our texture models on 8 texture images (D4, D6, D16, D21, D53, D68, D77 and D103)
from the Brodatz texture dataset. Acording to \citet{Lin-CVPR2006}, we can roughly classify
them into 4 different types, regular textures (D6, D21, D53, D77), near-regular textures (D16, D103),
irregular textures (D68) and stochastic textures (D4). The regular textures are simpler:
shallow models (such as mPoT, Gaussian RBM and ssRBM with tiled-convolutional weight sharing) are able to model them with high fidelity. 
However, the other textures (D4, D16, D68 and D103) remain challenging for shallow models. We show that deep models give better results.

\begin{figure*}[htbp!]
\centering
\includegraphics[scale=0.5]{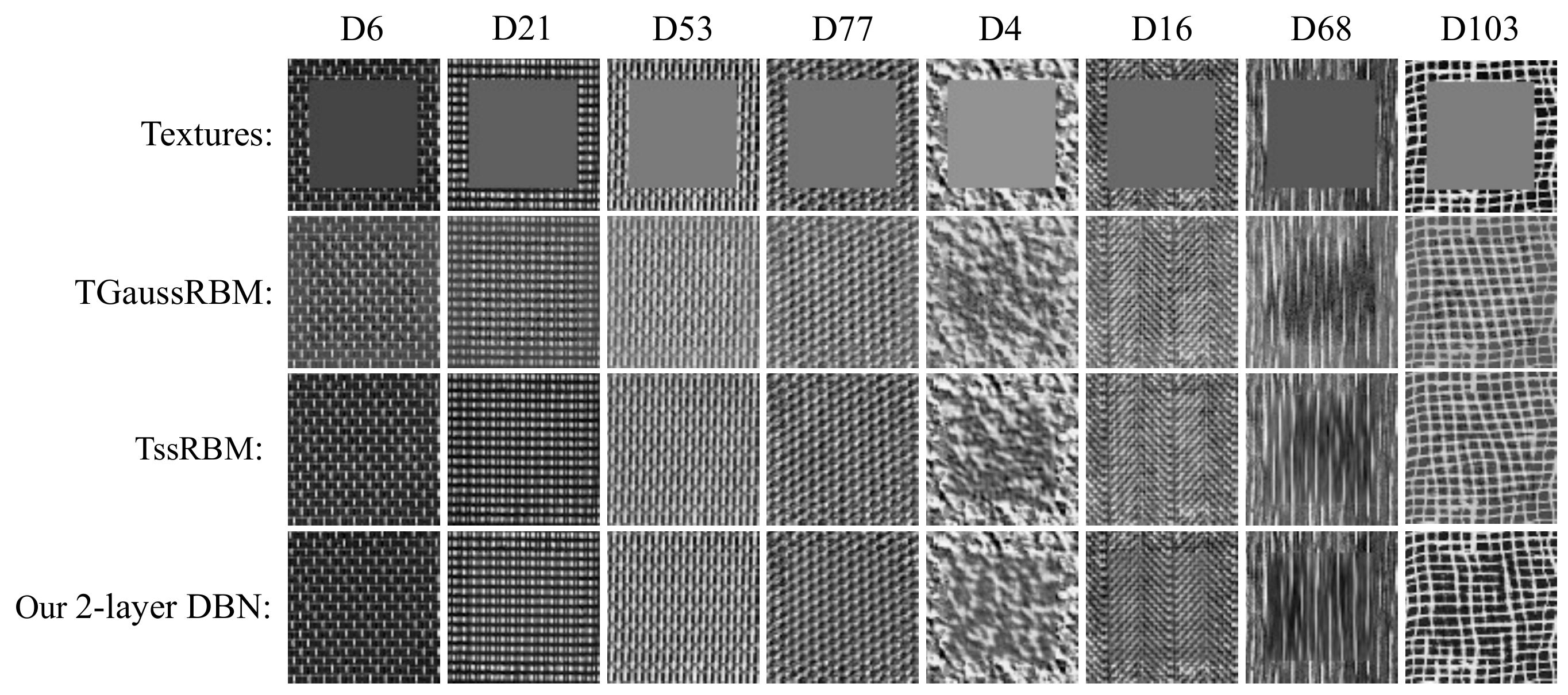}
\caption{\small Examples of texture inpainting for the models under consideration (rows) for different textures (columns).}
\label{fig:texture_inpaint}
\end{figure*}

\begin{table*}
\scriptsize
\begin{center}
\begin{tabular}{ccccc}
Synthesis                & D6              & D21             & D53             & D77 \\ \hline
Bi-FoE                   & 0.7573 $\pm$ 0.0594 & 0.8710 $\pm$ 0.0317 & 0.8266 $\pm$ 0.0869 & 0.6464 $\pm$ 0.0215 \\
TmPoT                    & 0.9329 $\pm$ 0.0356 & 0.8961 $\pm$ 0.0696 & 0.8527 $\pm$ 0.0559 & \bf{0.8699 $\pm$ 0.0080} \\
TPoT                     & 0.5641 $\pm$ 0.0916 & 0.7388 $\pm$ 0.1055 & 0.7583 $\pm$ 0.1082 & 0.6870 $\pm$ 0.0973 \\
T-GaussRBM               & 0.9301 $\pm$ 0.0207 & 0.8901 $\pm$ 0.0792 & 0.8485 $\pm$ 0.0606 & \bf{0.8663 $\pm$ 0.0084} \\
Multi-Tm (256)           & 0.9304 $\pm$ 0.0280 & 0.9346 $\pm$ 0.0205 & 0.9231 $\pm$ 0.0103 & 0.8610 $\pm$ 0.0096 \\ \hline
TssRBM                   & 0.9365 $\pm$ 0.0468 & \bf{0.9482 $\pm$ 0.0249} & \bf{0.9412 $\pm$ 0.0215} & 0.8410 $\pm$ 0.0121 \\
Our 2-layer DBM          & \bf{0.9516 $\pm$ 0.0164} & \bf{0.9465 $\pm$ 0.0322} & \bf{0.9499 $\pm$ 0.0264} & \bf{0.8638 $\pm$ 0.0161} \\ \hline
\\
Inpainting               & D6              & D21             & D53             & D77 \\ \hline
Efros\&Leung             & 0.8524 $\pm$ 0.0318 & 0.8566 $\pm$ 0.0344 & 0.8558 $\pm$ 0.0578 & 0.6012 $\pm$ 0.0760 \\
TmPoT                    & 0.8629 $\pm$ 0.0180 & 0.8741 $\pm$ 0.0116 & 0.8602 $\pm$ 0.0234 & \bf{0.7668 $\pm$ 0.0322} \\
TPoT                     & 0.8446 $\pm$ 0.0172 & 0.8609 $\pm$ 0.0275 & 0.8935 $\pm$ 0.0159 & 0.6379 $\pm$ 0.0373 \\
T-GaussRBM               & 0.8578 $\pm$ 0.0160 & 0.8662 $\pm$ 0.0185 & 0.8494 $\pm$ 0.0233 & \bf{0.7642 $\pm$ 0.0267} \\
Multi-Tm (256)           & 0.8452 $\pm$ 0.0173 & 0.8673 $\pm$ 0.0103 & 0.8554 $\pm$ 0.0284 & 0.7328 $\pm$ 0.0615 \\ \hline
TssRBM                   & \bf{0.8881 $\pm$ 0.0227} & \bf{0.9119 $\pm$ 0.0139} & \bf{0.9156 $\pm$ 0.0237} & \bf{0.7627 $\pm$ 0.0314} \\
Our 2-layer DBN          & \bf{0.8894 $\pm$ 0.0246} & \bf{0.9060 $\pm$ 0.0160} & \bf{0.9242 $\pm$ 0.0285} & \bf{0.7738 $\pm$ 0.0232} \\ 
\hline
\end{tabular}
\end{center}
\caption{\small A comparison of the one and two-layer TssRBM results with other models. All reported results other than the TssRBM-based results were taken from \citet{Kivinen2012} (including their Multi-Tm: a multiple texture model trained with 256 hidden units). The synthesis results are based on the TSS criterion
while the inpainting results are based on MSSIM-scores. In both cases larger numbers are better.}
\label{tab:simple_textures}
\end{table*}

\vspace*{-2mm}
\subsection{TssRBM texture modeling}
\vspace*{-2mm}

In this section, we compare the tiled-convolutional ssRBM  with other related models in the literature. We base our comparison on the results reported in \citet{Kivinen2012}. 
To provide a fair comparison, we follow the general experimental protocol established by \citet{Heess2009} and \citet{Kivinen2012}. Specifically, we rescaled the original 
$640 \times 640$ textures (all but D16) to either $480 \times 480$
(D4, D21 and D77) or $320 \times 320$ (D6, D53, D68 and D103). Each texture image was divided into a top half for training and a bottom half used
for testing. Then we report the performances of the TssRBM and our 2-layer TssRBM-based DBN 
on two tasks: texture synthesis and inpainting. All
models, in all experiments, are trained on $98 \times 98$ sized patches randomly cropped from the preprocessed training
texture images which are normalized to have zero mean and standard deviation of 1. We use a minibatch size of 64. 

The TssRBM is trained with FPCD~\citep{TielemanT2009}. For deep models,
we always pretrain the lower layer with one step CD and train the top layer with PCD (We find that in the higher
layer RMBs, the mixing of the negative phase Gibbs chain is relatively fast, so we use PCD). In both PCD and FPCD training processes, at the 
beginning of learning the persistent chains are initialized with noise and for some textures (especially
for those regular textures) restarting the Markov chains with a small possibility, like 0.01, seems advantageous to further promote mixing. After training, we aply our models for the following two task: texture synthesis and inpainting.

\begin{figure*}[htbp!]
  \centering
    \vspace{-3mm}    
    \includegraphics[scale=0.3]{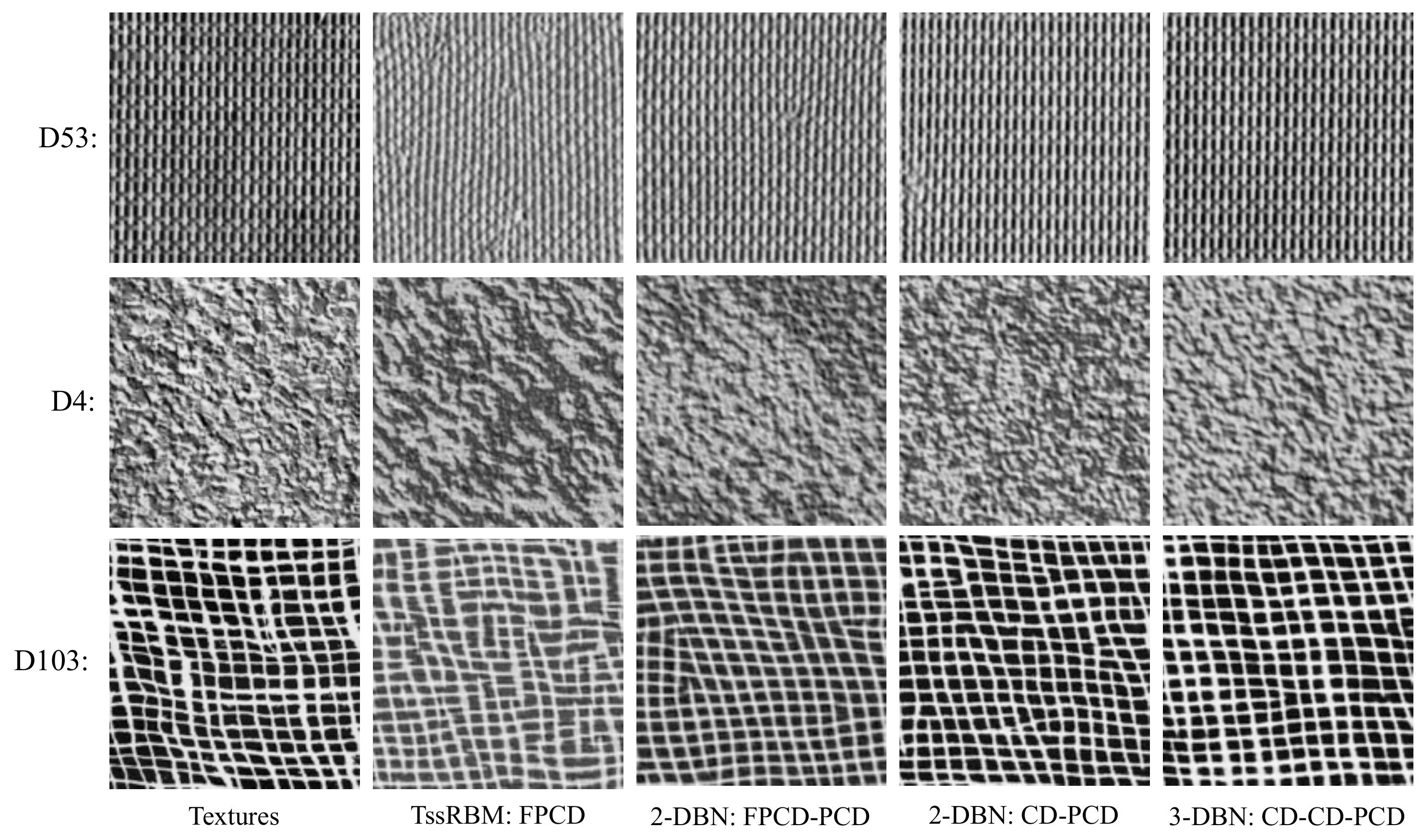} \includegraphics[scale=0.38]{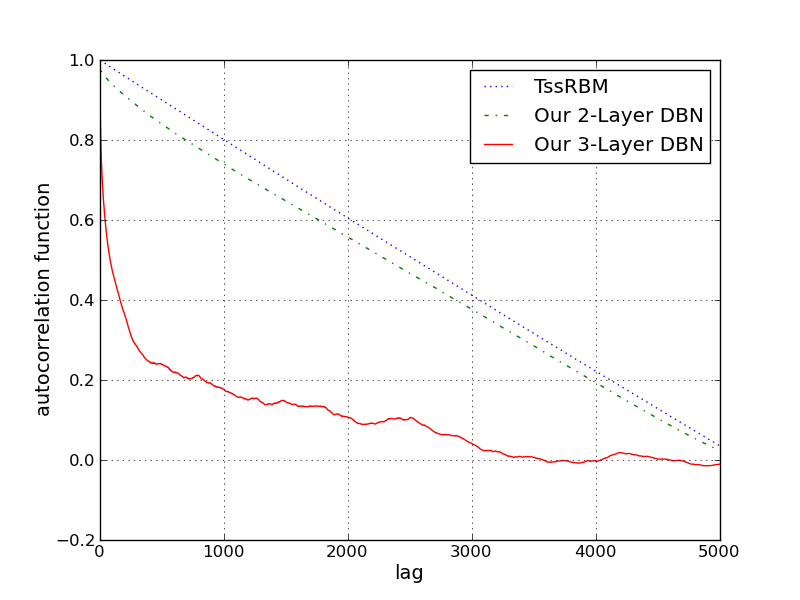} 
    \vspace{-4mm}    
\caption{\small {\bf LEFT:} Synthesized texture D53, D4 and D103 at \emph{full resolution}. The training algorihtms are shown in the layer-order, e.g. 3-DBN: CD-CD-PCD denotes a 3-layer DBN trained with CD for the first two layers and with PCD for the upppermost layer. Both depth and the choice of inductive bias have a significant impact on the quality of the model. {\bf RIGHT:} The autocorrelation spectrum of Monte Carlo Markov Chain samples of the texture D103 for our one, two and three-layer models. All layers are trained with CD, except the uppermost which is trained with PCD (TssRBM trained with FPCD).}
\label{fig:highres_textures}
    \vspace{-3mm}    
\end{figure*}

\paragraph{Texture Synthesis:}
For this task we generate unconstrained samples from our models by the usual DBN generative procedure,
with Gibbs sampling in the top-level RBM, followed (in the case of deep models)
by stochastic projection (except for the visible units, as usual, and except for the slab units, where we take the expectation) in image space. 
Following~\citet{Kivinen2012}, after a large number of ``burn-in'' samples, we collected 128 samples of size $120
\times 120$ for both the 1-layer and 2-layer models. A quantitative measure of the quality of the samples 
is provided by the
Texture Similarity Score (TSS)~\citep{Heess2009}, comparing each generated sample
with the test patches from the test region of the
image. For a sample $s$ and test texture $x$, the TSS is given by the
maximum normalized cross correlation (NCC):
\begin{equation}
TSS(s,x) = \max \left\{ \frac{x_{1}^Ts}{\|x_{1}\| \|s\|},\dots,\frac{x_{L}^Ts}{\|x_{L}\| \|s\|} \right\},
\end{equation}
where $x_{i}$ denotes patch $i$ within the test region of the image and $L$
is the number of possible unique patches in the test region. A patch (and
sample) of size $19 \times 19$ was used to compute the score. We only use
TSS for those regular textures (D6, D21, D53,
D77). Fig. \ref{fig:texture_synthesis} compares images of textures
synthesized by some of the methods under consideration. Table
\ref{tab:simple_textures} provides a quantitative comparison based on the
TSS and shows that the TssRBM-based models are competitive with these other
probabilistic models of texture.

\paragraph{Inpainting:}
The inpainting (constrained texture synthesis) task requires the models to generate a texture
which is consistent with a given boundary. Following~\citet{Kivinen2012}, we randomly cut
$76 \times 76$ texture patches from the test texture images and set the center ($54
\times 54$) to zeros. The resulting images as the inpainting frames were fed to our models.
The inpainting was done by running 500 Gibbs sampling iterations in our models while the
border was held fixed. The number of inpainting frames was 20 for each texture, and the inpainting 
were each done with 5 different random seeds, making it a total of 100 
inpaintings for each model and each texture. The quality of the inpainting was evaluated 
using the mean structural similarity index (MSSIM)~\citep{Wang2004} that compares 
the inpainted region and the ground truth. Fig. \ref{fig:texture_inpaint} compares the texture results of some of the methods under consideration. Table \ref{tab:simple_textures} provides the quantitative MSSIM comparison against other similar models. Here again, the TssRBM-based models are fairly competitive with these other probabilistic models of texture. 




\begin{figure*}[htbp]
\label{fig:8_texture}
\centering
\includegraphics[scale=0.35]{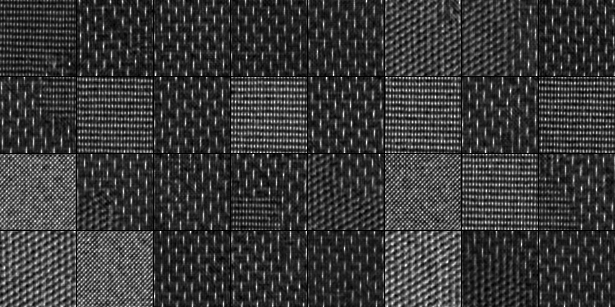} \hspace{1mm}
\includegraphics[scale=0.35]{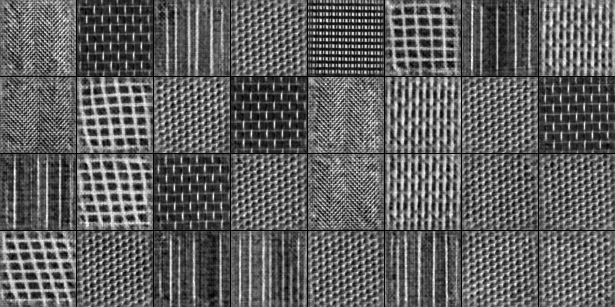}
\caption{{\bf LEFT:} Multi-texture samples generated by the TssRBM model. {\bf RIGHT:} Multi-texture samples generated by our 3-layer DBN.}
\label{fig:multi_texture}
\end{figure*}

\vspace*{-2mm}
\subsection{Experiments II: Exploring High-Resolution Textures}
\vspace*{-2mm}

To further explore the generative power of the DBN models,
we move to a more challenging task, specifically, modeling high-resolution textures while keeping the first layer structure unchanged:
the same number of filters, the same size ($11\times11$) of the receptive
fields and the same size ($98\times98$) of the training patches.  This implies that the first layer will face a much more challenging 
learning task. We show that by adding more hidden layers these difficult tasks are handled very well. We add two more hidden layers 
to the first layer. That gives us three layer DBNs. There are 128 filters with convolutional weight sharing in both of these two layers.  
Due to the limited sapce, we only show the results of texture D53, D4 and D103. The other 5 textures yield a similar pattern of results. While the quantitative measures used in the previous experiments are useful to extablish an objective comparison between methods, we feel that they are rather imperfect measures of the quality of the texture model and therefore in this section we forgo these measures in favour of simply presenting texture synthesis results for visual inspection. Fig. \ref{fig:highres_textures} (right) illustrates the impact of both depth and the inductive bias (FPCD versus CD training) in training TssRBM-based models of texture.

\paragraph{Depth helps mixing.}
One key aspect that might help to explain the improvements in the models is
that as the model gets deeper the mixing rate of the negative phace Gibbs
chain improves, as already demonstrated and argued
in~\citet{Bengio-arxiv-mixing-2012}. Improved mixing of the Gibbs chain
improves the performance of training methods such as PCD that rely on it
for the estimation of negative phase statistics. It also helps the generation
of the samples shown. To demonstrate the
improvement in mixing with depth, we assess the mixing rate of three models
(one, two and three layer model) trained on D103 via the autocorrelation spectrum. 
After training, we run a Markov chain in all of three models and plot the
autocorrelation spectrum in Fig. \ref{fig:highres_textures} (left). As seen in the
figure, mixing become very fast in the three-layer model, i.e., samples
at some distance in the chain are less correlated with each other.

\paragraph{CD pretraining vs. PCD and FPCD pretraining.}
We find that pretraining the lower layers with CD-1 results in better DBN texture models. More specifically,
worse results were obtained by PCD pre-training, then FPCD pre-training, and substantially better with CD, as can e.g. be
seen in Figure~\ref{fig:highres_textures} (right). This is consistent with the claims made in~\citet{Hinton-ipam2012} regarding the
advantage of CD vs PCD. It is also consistent with the results in \citet{LeRoux-Bengio-2008}, which show that maximum
likelihood training of the lower layers of a DBN is sub-optimal, and that assuming a high-capacity top layer, the optimal
way to train the first layer would be to minimize the KL divergence between the visible units and the stochastic
one-step reconstruction, something much closer to what CD does than what PCD does. Another hypothesis is that CD helps here because it makes sure to
extract {\em good features that preserve the input information}, without the constraint that the lower level RBMs
do a good job (of avoid spurious modes) far from the training samples. Instead for the top-level RBM, which is
used to sample from the model, it is important to use a good approximation to maximum likelihood training.

\vspace*{-2mm}
\subsection{Learning with Multiple Textures}
\label{sec:multi_texture}
\vspace*{-2mm}

In this section, we try to assess the power of our deep models by using not only high-resolution
texture images but also multiple heterogeneous textures. We train a three layers model on all 8 textures. The first layer of our DBN is a TssRBM
with 96 filters. The second layer is our new RBM variant introduced in Sec. \ref{sec:newRBM} with 256 filters and receptive fields of size $2 \times 2$. The third layer is a convolutional binary RBM with 256 filters and receptive fields of size $2 \times 2$. We compare our DBN with a one layer model (TssRBM with 128 filters).
After training, we generate samples from both models and show the results in Fig. \ref{fig:multi_texture}. 
We can see that the single layer TssRBM only models the high frequency structure in the training data. On the other hand,
the deep model seems to capture much of the 8 textures that occur in the training set. There are 7 different textures apparent in these 32 samples. We are only missing samples of D4, which 
is a stochastic texture and hard to capture, particularly when most of the training data are highly structured
images.  \citet{Kivinen2012} also trained Gaussian RBM with tiled-convolution weight sharing on
multiple textures with labels. The labels can help their model to pick different filters for different 
textures and thus make the learning problem much easier.  

\section{Conclusions}

In this paper, we apply the ssRBM with tiled-convolution weight sharing on
texture modeling task.  We show that not only is the ssRBM competitive as a
single layer model of texture, but that, by being amenable to CD training,
it it well suited to being incorporated into even more effective deep
models of texture. Interestingly, we find that CD training of lower layers
yields better models, and that mixing is better in deeper layers.  Our
integration of the ssRBM into a DBN necessitated the development of a novel
RBM with a spike-and-slab visible layer and a binary latent layer.  Finally
we show our new ssRBM-based DBN is capable of modeling multiple
high-resolution textures.



\small
\bibliography{strings,strings-shorter,ml,aigaion}
\bibliographystyle{natbib}

\end{document}